\documentclass[nohyperref]{article}
\usepackage{microtype}
\usepackage{graphicx}
\usepackage{subfigure}
\usepackage{booktabs} 
\usepackage{hyperref}

\usepackage[accepted]{icml2023}
\usepackage{amsmath}
\usepackage{amssymb}
\usepackage{mathtools}
\usepackage{amsthm}
\usepackage{comment}
\usepackage[capitalize,noabbrev]{cleveref}

\theoremstyle{plain}

\theoremstyle{definition}

\theoremstyle{remark}

\usepackage[textsize=tiny]{todonotes}
\usepackage{makecell,array}
\usepackage{enumitem}

\icmltitlerunning{Learning Diverse Features in Vision Transformers for Improved Generalization}

\begin{document}

\twocolumn[
\icmltitle{Learning Diverse Features in Vision Transformers\\for Improved Generalization}



\icmlsetsymbol{equal}{*}

\begin{icmlauthorlist}
\icmlauthor{Armand Mihai Nicolicioiu}{idiap}
\icmlauthor{Andrei Liviu Nicolicioiu}{mila,udem}
\icmlauthor{Bogdan Alexe}{unibuc,ismma}
\icmlauthor{Damien Teney}{idiap}
\end{icmlauthorlist}

\icmlaffiliation{idiap}{Idiap Research Institute, Martigny, Switzerland}

\icmlaffiliation{mila}{Mila, Montreal, Canada}
\icmlaffiliation{udem}{University of Montreal, Canada}
\icmlaffiliation{unibuc}{University of Bucharest, Romania}

\icmlaffiliation{ismma}{Gheorghe Mihoc - Caius Iacob Institute of Mathematical Statistics and Applied Mathematics of the Romanian Academy}

\icmlcorrespondingauthor{Armand Mihai Nicolicioiu}{armand.nicolicioiu@gmail.com}




\icmlkeywords{vision transformers, OOD, out-of-distribution generalization, distribution shifts}

\vskip 0.3in
]



\printAffiliationsAndNotice{}  

\begin{abstract}
Deep learning models often rely only on a small set of features even when there is a rich set of predictive signals in the  training data.
This makes models brittle and sensitive to distribution shifts.

In this work, we first examine vision transformers (ViTs) and find that they tend to extract robust and spurious features with distinct attention heads.
As a result of this modularity, their performance under distribution shifts can be significantly improved at test time by pruning heads corresponding to spurious features, which we demonstrate using an ``oracle selection'' on validation data.

Second, we propose a method to further enhance the diversity and complementarity of the learned features by encouraging orthogonality of the attention heads' input gradients.
We observe improved out-of-distribution performance on diagnostic benchmarks (MNIST-CIFAR, Waterbirds) as a consequence of the enhanced diversity of features and the pruning of undesirable heads.
\end{abstract}

\section{Introduction}
\label{introduction}

State-of-the-art models in machine learning show their limits when faced with distribution shifts.
Even though they can perform remarkably well when training and test data are drawn from the same distribution, their predictive performance can degrade dramatically otherwise.
The reason is that features that are predictive in the training data may be spurious and misleading at test time.
Out-of-distribution (OOD) generalization is the capability of a model to maintain its predictive performance in the face of such shifts.

Prior work has shown that deep learning models often rely only on a small set of predictive features~\cite{geirhos2020shortcut}.
If any of these features are spurious and affected by a distribution shift, chances are high that a model's performance will be affected.
A recent line of work seeks to increase the diversity of the learned features, either as an objective when training a predictive model~
\cite{ross2020ensembles,teney2022evading,lee2022diversify} or as a step prior to the application of invariance-learning methods~\cite{chen2023towards,zhang2022rich}.

This paper focuses on computer vision tasks and vision transformers (ViT) \cite{dosovitskiy2020image}.
We apply a regularizer based on input gradients~\cite{ross2020ensembles,teney2022evading} to a ViTs' attention heads to diversify the features learned across these heads.
This encourages different parts of the model to rely on different aspects of the data and to discover additional predictive patterns.
In contrast to methods that diversify functional behaviour in prediction space~\cite{lee2022diversify,chen2023project}, our approach operates in feature space and does not require any OOD data (even unlabeled) during training.

\begin{figure*}
    \centering
    \includegraphics[width=0.88\textwidth]{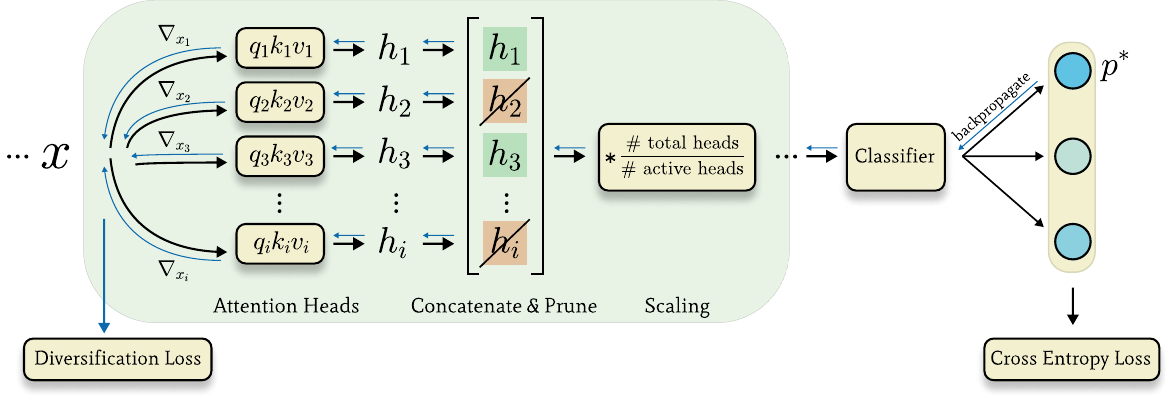}
    \caption{\textbf{Overview of the proposed DiverseViT method.} We depict a single-layer multi-head attention model.
    }
    \label{fig/diversevit-overview}
\end{figure*}

This paper presents early experiments on standard OOD benchmarks (MNIST-CIFAR, Waterbirds).
First, we find that \textbf{ViTs already have an inherent property for modularity}: their attention heads rely each on different features, such that they can be pruned selectively to discard spurious ones and improve generalization.
Second, we show that \textbf{the proposed regularizer can further increase the diversity and complementarity of the learned features}.
Our method, \textit{DiverseViT} (see Figure \ref{fig/diversevit-overview}), leads to improvements in a standard OOD evaluation setting, and even more so when we allow pruning attention heads at test time using for selection the highest accuracy obtained on a labeled validation set from the target distribution.%
\footnote{This setting provides an upper bound on the performance achievable with ideal model selection heuristics.}

\renewcommand{\thefootnote}{}
\footnotetext{Code available at \href{https://github.com/ArmandNM/diverse-vit}{https://github.com/ArmandNM/diverse-vit}.}

\paragraph{Summary of contributions.}
\setlist{nolistsep,leftmargin=*}
\begin{itemize}[topsep=-4pt,itemsep=2pt,labelindent=*,labelsep=7pt,leftmargin=*]

\item We evaluate off-the-shelf ViTs on diagnostic OOD benchmarks and find inherent modularity in their representations, such that OOD generalization can be improved by pruning the attention heads that rely on spurious features.

\item We propose a simple regularizer to increase the diversity of features learned by ViTs.

\item We evaluate models trained with our method (DiverseViT) and observe increased diversity and complementarity of the learned features. They show better OOD performance in standard evaluations, and yet much higher performance when pruning specific attention heads at test time.
\end{itemize}



\section{Related Work}
\label{related work}

\paragraph{Diversity of solutions in machine learning.}
A range of methods have been proposed to train models that are diverse in properties such as 
OOD generalization~\cite{lee2022diversify,teney2022predicting},
interpretability~\cite{chen2023understanding,semenova2022existence}, or fairness. 
These diversification methods are relevant in cases of {underspecification}~\citep{d2020underspecification} when the standard ERM objective (empirical risk minimization) does not constrain the solution space to a unique one.

\paragraph{Increasing diversity.}
Most diversification methods train a set of multiple \emph{entire} models. By contrast, our approach seeks to increase diversity {within} a single transformer.
Existing methods train multiple models in parallel or sequentially.
They encourage diversity in \textbf{feature space}~\citep{heljakka2022representational,yashima2022feature}, 
\textbf{prediction space}~\citep{pagliardini2022agree,lee2022diversify}, or \textbf{gradient space}~\citep{ross2018learning,ross2020ensembles,teney2022evading,teney2022predicting}.
Our method applies a gradient-based approach similar to that of \citet{ross2020ensembles,teney2022evading} to the attention heads of ViTs.

\paragraph{Vision transformers (ViTs).}
We focus on ViTs because they archieve state-of-the-art performance for multiple tasks in computer vision~\cite{khan2022transformers}.
Multiple works have sought to understand the features learned by these models~\cite{bhojanapalli2021understanding,naseer2021intriguing,zhou2022understanding}.
Our findings are complementary. 
We also seek to nudge the (generally beneficial) inductive biases of ViTs.
In particular, we seek to overcome the general ``simplicity bias'' of deep learning models~\cite{shah2020pitfalls}. 


\section{Proposed method}
\label{proposedMethod}

We describe a simple method to diversify the features learned by ViTs trained for image classification.
We propose a regularizer that encourages orthogonality of the input gradients corresponding to their attention heads.
We show empirically that this provides a better inherent robustness to distribution shifts.
Moreover, this allows further improvements in OOD performance with test-time pruning of the attention heads that correspond to spurious features, while retaining those necessary for robust classification.

\subsection{Background: ViTs and attention heads}

The input image to a ViT is partitioned into a grid of small patches. A sequence of tokens is formed by combining the patches with positional embeddings. The main operation to aggregate a sequence of tokens is the multi-head self-attention, defined as:
\setlength{\abovedisplayskip}{3.3pt}
\setlength{\belowdisplayskip}{3.3pt}
\setlength{\abovedisplayshortskip}{3.3pt}
\setlength{\belowdisplayshortskip}{3.3pt}
\begin{equation}
    h_i = \text{softmax} \Bigg( \frac{x W_q (x W_k )^T}{\sqrt{D}}\Bigg) x W_v 
\end{equation}
where $x \in \mathbb{R}^{N \times D} $ represents the set of $N$ input tokens of the self-attention layer and $W_q, W_k, W_v \in \mathbb{R}^{D \times D} $ are learnable parameters of the layer.

The output of all heads are concatenated and the result is projected with a linear mapping:
\begin{equation}
\label{eq:self-att-concat}
    y = [h_1, ..., h_H] W_o
\end{equation}

with $W_o \in \mathbb{R}^{D \times H*D}$ learnable parameters.
\subsection{Encouraging feature diversity with input gradients}
\label{diversification}
Our diversification method is a regularizer term added to the optimization objective when training a ViT for a supervised task with standard ERM.
This approach increases the diversity of features {within} a single transformer, which contrasts with many diversification methods that train multiple entire models.
The motivation for our approach is (1)~its lower computational cost and (2)~leveraging the existing inductive biases of vision transformers to avoid the type of ``adversarial'' solutions to the gradient-based regularizer that were described in~\citet{teney2022predicting}.

\paragraph{Input gradients.}
To determine how much each dimension of a feature vector contributes to the prediction of the model, we look at the gradient of the prediction with respect to this vector.
Concretely, we compute the gradient of the top predicted score $p^*$ with respect to the input $x$:
\begin{equation}
\label{eq:grad_top_pred}
    \nabla_x = \frac{\partial p^*}{\partial x} \in \mathbb{R}^{N \times D}.
\end{equation}
\paragraph{Influence of each head.} The outputs of all attention heads are concatenated and projected (Equation \ref{eq:self-att-concat}), so $\nabla_x$ considers all attention heads' effects simultaneously. As we are interested in diversifying the effect of each individual head, we want to capture their individual contributions. For this, we backpropagate the gradient of the top prediction (Equation \ref{eq:grad_top_pred}) $H$ times, each time through a single element $h_i$ of Equation \ref{eq:self-att-concat} while ignoring the rest. We obtain a set of $H$ input gradient:
\begin{equation}
 \Big\{ \nabla_{x_i} \in \mathbb{R}^{N \times D} \mid i \in \{1 \cdots H\} \Big\}.
\end{equation}
Each element in this set represents the importance of the input features to the trop prediction for a specific head, with 
\mbox{$\sum_{i=1}^{H}\nabla_{x_i}=\nabla_x$}.

\paragraph{Diversity regularizer.} To promote diversity across the heads, we define an orthogonality regularizer over the input gradients.
We first compute the cosine similarity between the $n$th token in heads $i$ and $j$ by normalizing over the channel dimension $D$ and taking the dot product between all tokens:
\begin{equation}
    c_{n,i,j} = \nabla_{x_{i,n}}^T \nabla_{x_{j,n}} \in \mathbb{R}.
\end{equation}
The orthogonality regularizer is then defined as the average squared similarity across all tokens and pairs of heads:
\begin{equation}
    \label{eq:loss_ig}
    \mathcal{L}_{Div} = \frac{1}{N} \sum_{i\neq j}  \sum_{n=1}^{N} c_{n,i,j} ^ 2.
\end{equation}
The regularizer, weighted by a hyperparameter $\lambda$, is added to the standard cross-entropy classification loss to form the complete training objective:
\begin{equation}
    \mathcal{L} ~=~ \mathcal{L}_{ERM} + \lambda \,\mathcal{L}_{Div}.
\end{equation}

\subsection{Pruning attention heads}
\label{headPruning}

Different attention heads of a transformer can attend to different features which can be more or less robust (i.e. useful across distribution shifts) or spurious.
Therefore, using only a subset of heads is a simple technique for ignoring undesirable features and improving OOD performance.

To prune a subset of the heads,
we multiply their output $h_i$ by zero in Equation \ref{eq:self-att-concat}. To compensate for the missing heads, we scale the remaining ones to remain in the same range after the projection. This can be done at test time with no need for further adaptation of the model.

\section{Experiments}
\label{experiments}

We present experiments on two popular diagnostic datasets: MNIST-CIFAR and Waterbirds. See Appendix~\ref{appendixDatasets} for details.
Our experiments answer the following questions:
\begin{enumerate}
    \item Does diversifying the learned features of the self-attention heads lead to better OOD performance compared with ERM training? \textbf{Yes.}
    \item Can we improve generalization by pruning heads associated with spurious features with a standard ERM-trained ViT (i.e. without our diversification method)? \textbf{Yes.}
    \item Does our diversification method amplify the distinction between spurious and robust features, improving post-pruning performance even more? \textbf{Yes.}
\end{enumerate}


The \textbf{baseline} experiment ({\sc ViT+ERM}) trains a ViT with ERM.
In both datasets used, a spurious feature is strongly correlated with the label during training. This correlation is reversed at test time. A model that relies on this spurious feature will therefore perform poorly at test time.

In the \textbf{diversification} experiment ({\sc ViT+Div}), we add our diversity regularizer to the training objective. Without any changes in the architecture, we obtain better generalization, possibly as an ensemble effect of a more diverse set of features captured collectively by all attention heads.

For the \textbf{head selection} experiments ({\sc Sel}), we perform inference at test time using a {single attention head}.
We select the head {with the highest accuracy on the OOD validation data}.
This therefore requires access to labeled OOD examples. The pruning is a test-time procedure and requires no further training of the model.
In all experiments, we select hyperparameters for highest OOD validation accuracy.

\begin{figure*}
    \centering
    \includegraphics[width=0.93\textwidth]{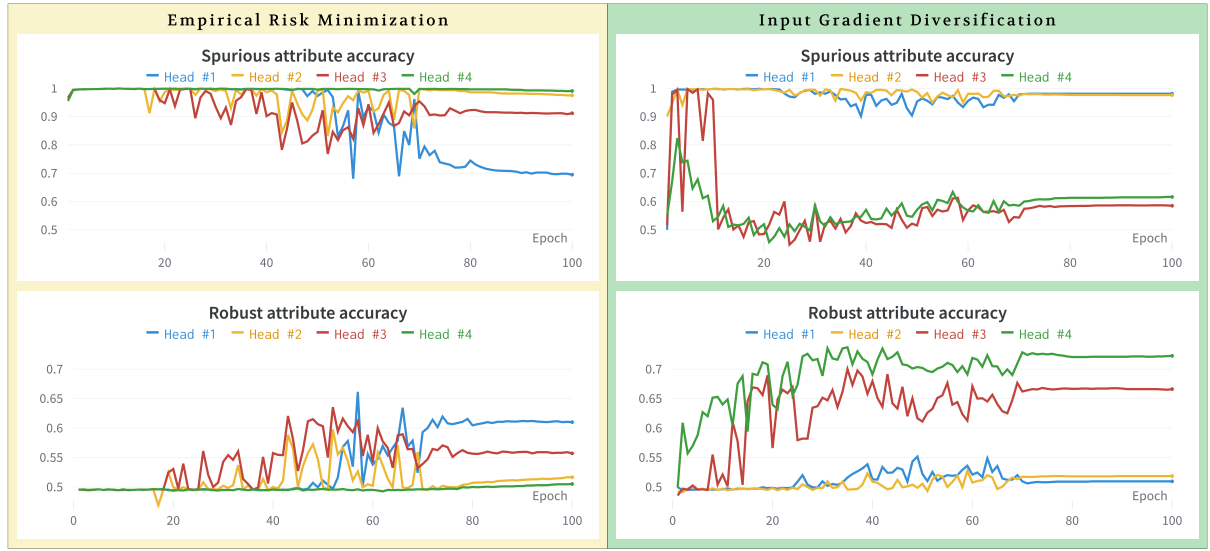}
    \caption{\textbf{Per-head performance comparison between standard ERM and our Diversification method on MNIST-CIFAR.} We observe an inherent modularity of ViT's attention heads, such that the heads predicting well on the robust attribute are predicting poorly on the spurious one, and vice-versa. With standard ERM, the gap between the two attributes predictions is not as clear for most heads, indicating a higher overlap in the information captured by each head. With the proposed diversification method, we observe that most heads predict either, but not both of the robust and spurious features. This shows a high level of specialization, which is desirable since this allows pruning undesirable heads without losing information relevant to robust predictions.
    }
    \label{fig/per-head-performance}
    \vspace{-0.1in}
\end{figure*}


\paragraph{Main results.} In Tables~\ref{table:results-mnist-cifar}--\ref{table:results-waterbirds} we report accuracy on both ID (in-domain) and OOD (out-of-distribution) test sets.
We observe that the ERM baseline already exhibits a degree of separation between spurious and robust features among the self-attention heads, thus allowing the head selection~(\textit{ViT+ERM+Sel}) to improve the OOD accuracy.
The diversification method~(\textit{ViT+Div}) is effective on its own (i.e. even without head selection). This indicates a benefit from learning diverse features.
The combination of diverse features and proper head selections~(\textit{ViT+Div+Sel}) is especially powerful, leading to our best result.

\begin{table}[h]
    \vskip -0.05in
    \caption{Results on MNIST-CIFAR.}
    \label{table:results-mnist-cifar}
    \vskip -0.05in
    \begin{center}
    \begin{small}
    \begin{sc}
    \begin{tabular}{lccr}
        \toprule
        \multicolumn{1}{p{0.2\columnwidth}}{Method} & \multicolumn{1}{m{0.29\columnwidth}}{\centering ID Accuracy} & \multicolumn{1}{m{0.29\columnwidth}}{\centering OOD Accuracy} \\
        \midrule
        ViT+ERM    & 88.80 $\pm$ 0.12& 56.87 $\pm$ 4.30 \\
        ViT+Div & 88.40 $\pm$ 0.10 & 62.26 $\pm$ 1.80 \\
        ViT+ERM+Sel    & \textbf{90.33} $\pm$ 0.19& 64.40 $\pm$ 2.80 \\
        ViT+Div+Sel    & 89.86 $\pm$ 1.10& \textbf{70.08} $\pm$ 3.15 \\
        \bottomrule
        \end{tabular}
    \end{sc}
    \end{small}
    \end{center}
    \vskip -0.15in
\end{table}

\paragraph{Understanding the learned features}
To gain insight into the learned features, we evaluate the models on a balanced split of MNIST-CIFAR with no correlation between the robust and the spurious attributes. We measure the correlation between the model's predictions and either the robust or the spurious attribute. 
How well each attribute can be predicted with each head indicates how much of each attribute is captured by each head.
Figure~\ref{fig/per-head-performance} shows that diversification leads to a higher level of specialization of the heads. This is advantageous, allowing us to keep only the heads containing information relevant to robust predictions.


\begin{table}[t]
    \vskip -0.05in
    \caption{Results on Waterbirds.}
    \label{table:results-waterbirds}
    \vskip -0.05in
    \begin{center}
    \begin{small}
    \begin{sc}
    \begin{tabular}{lccr}
        \toprule
        \multicolumn{1}{p{0.2\columnwidth}}{Method} & \multicolumn{1}{m{0.29\columnwidth}}{\centering ID Accuracy} & \multicolumn{1}{m{0.29\columnwidth}}{\centering OOD Accuracy} \\
        \midrule
        ViT+ERM    & 96.55 $\pm$ 0.22& 83.37 $\pm$ 0.44 \\
        ViT+Div & 96.99 $\pm$ 0.11& 83.87 $\pm$ 0.79 \\
        ViT+ERM+Sel    & 96.50 $\pm$ 0.58& 85.70 $\pm$ 1.64 \\
        ViT+Div+Sel    & \textbf{96.99} $\pm$ 0.12& \textbf{87.96} $\pm$ 0.14 \\
        \bottomrule
        \end{tabular}
    \end{sc}
    \end{small}
    \end{center}
    \vskip -0.15in
\end{table}

\section{Conclusions and future work}
\label{conclusion}

We have shown that ViT have an inherent tendency to capture distinct features in their attention heads, and that this property can be improved with a simple  regularizer.
This directly improves the robustness of ViTs on several diagnostic benchmarks for out-of-distribution generalization, without changing the architecture. These improvements come ``for free'' as an ensembling effect from the increased diversity of the learned features.

We also empirically showed that these diverse features have little overlap and are complementary. Therefore, pruning selected attention heads at test time is an effective technique to improve OOD performance, using some information that can identify which heads correspond to spurious features (e.g. an OOD validation set).

\paragraph{Future work.}
Our methods should be evaluated on larger-scale real-world datasets in vision, but also language and reinforcement learning. The head pruning procedure was evaluated using labeled OOD data, which is only meant to provide an upper bound on the performance achievable in an ideal setting.
The approach should be evaluated with the recent heuristics proposed for OOD model selection~\cite{baek2022agreement,deng2023confidence,garg2022leveraging,liu2023neuron,lu2022towards}.
Other options include various forms of human feedback and unsupervised objectives to enable test-time adaptation.

\section*{Acknowledgements}
Bogdan Alexe was funded by UEFISCDI under Project EEA-RO-2018-0496. Damien Teney was partially supported by an Amazon Research Award (ARA).

\bibliography{bibliography}
\bibliographystyle{icml2023}

\clearpage
\appendix
\onecolumn

\section{Datasets}
\label{appendixDatasets}

\paragraph{MNIST-CIFAR}
The MNIST-CIFAR dataset \cite{shah2020pitfalls} contains collages of an image from MNIST (digits $0$ and $1$) and an image from CIFAR (a car or a truck) vertically concatenated. The true label comes from the vehicle type, and the digit is a spurious attribute highly correlated with the label during training. In our experiments, 90\% of examples containing the digit $0$ pair it with a car, and 90\% of examples containing the digit $1$ pair it with a truck. During the evaluation, this correlation no longer holds, and the images are evenly distributed.

\paragraph{Waterbirds}
The Waterbirds dataset \cite{sagawa2019distributionally} is constructed by adding a segmented image of a bird from the Caltech-UCSD Birds dataset over a background image from the Places dataset. The true label is the bird type (waterbird or landbird) and the spurious attribute is the background (water or land). In the training data, there are 3498 waterbirds on a water background and 1057 landbirds on a land background, while the minority groups contain only 184 waterbirds on a land background and 56 landbirds on a water background.

\begin{figure}[h]
    \centering
    \includegraphics[width=0.45\textwidth]{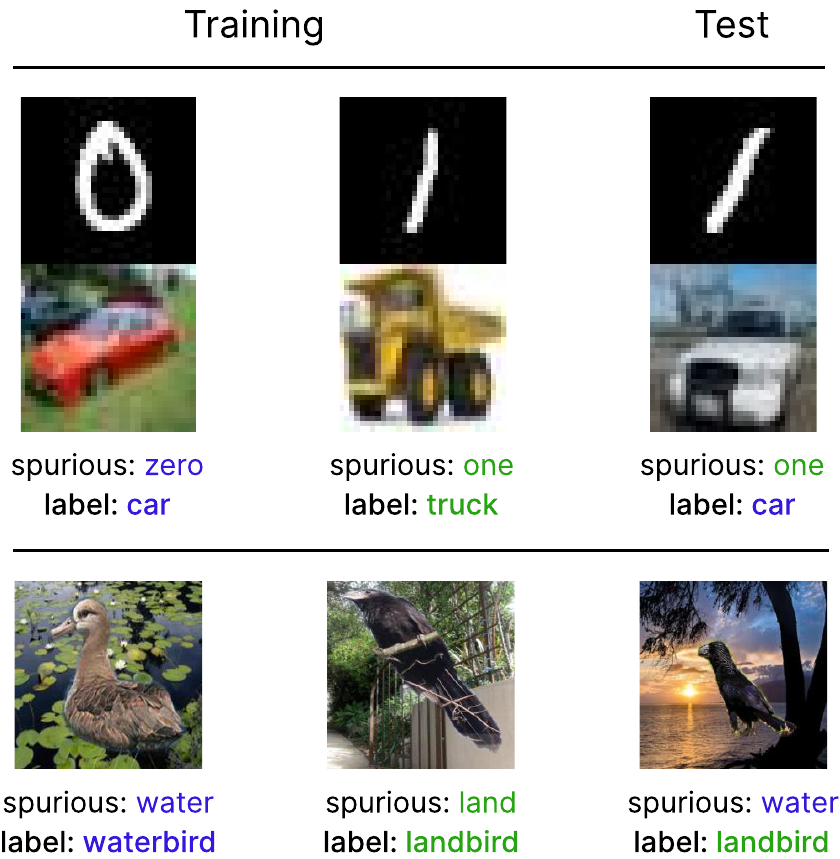}
    \caption{Samples from MNIST-CIFAR dataset (top) and Waterbirds (bottom).}
    \label{fig:enter-label}
\end{figure}

\end{document}